# Deep Learning Reveals Patterns of Diverse and Changing Sentiments Towards COVID-19 Vaccines Based on 11 Million Tweets

Hanyin Wang[1], Meghan R. Hutch[1], Yikuan Li[1], Adrienne S. Kline, PhD[1], Sebastian Otero[2], Leena B. Mithal, MD[2], Emily S. Miller, MD[3], Andrew Naidech, MD[4], Yuan Luo, PhD[1]

[1]Department of Preventive Medicine, Feinberg School of Medicine, Northwestern University, Chicago, IL, USA

[2]Ann & Robert H. Lurie Children's Hospital of Chicago and Department of Pediatrics, Feinberg School of Medicine, Northwestern University, Chicago, IL USA

[3]Department of Obstetrics & Gynecology, Northwestern Medicine, Chicago, IL, USA

[4]Department of Neurology, Northwestern Medicine, Chicago, IL, USA

Correspondence to: Yuan Luo, PhD, Yuan.Luo@Northwestern.edu, 750 N. Lake Shore Drive, 11-189, Chicago, IL, 60611

Abstract

*Background*

Vaccines have long been a crucial component of pandemic responses. Over 12 billion doses of COVID-19 vaccines have been administered at the time of writing. However, public perceptions of vaccines have been complex. We sought to use social media to understand the evolving perceptions of COVID-19 vaccines.

*Methods*

We analyzed COVID-19 vaccine-related tweets extracted from Twitter. After annotation, we finetuned a deep learning classifier using a state-of-the-art model, XLNet, to detect each tweet's sentiment automatically. We employed validated methods to extract the users' race or ethnicity, gender, age, and geographical locations from user profiles. Content specific to vaccination in pregnancy was isolated for separate analysis. Incorporating multiple data sources, we assessed the sentiment patterns among subpopulations and juxtaposed them against vaccine uptake data to unravel their interactive patterns.

*Findings*

11,211,672 COVID-19 vaccine-related tweets corresponding to 2,203,681 users over two years were analyzed. The finetuned model for sentiment classification yielded an accuracy of 0·92 against the hand-annotated testing set. Users from various demographic groups demonstrated distinguishable patterns in sentiments towards COVID-19 vaccines. User sentiments became more positive over time, upon which we observed subsequent upswing in the population-level vaccine uptake. Surrounding dates where positive sentiments crest, we detected encouraging news or events regarding vaccine development and distribution. Positive sentiments in pregnancy-related tweets demonstrated a delayed pattern compared with trends in general population, with postponed vaccine uptake trends.

*Interpretation*

Distinctive patterns across subpopulations suggest the need of tailored strategies. Global news and events profoundly involved in shaping users' thoughts on social media. Populations with additional concerns, such as pregnancy, demonstrated more substantial hesitancy since lack of timely recommendations. Feature analysis revealed hesitancies of various subpopulations stemmed from clinical trial logics, risks and complications, and urgency of scientific evidence. The findings provided evidence-based strategies for improving vaccine promotion in future pandemics.

Research in context

*Evidence before this study*

We searched PubMed for articles published in English between December 1, 2019 to March 1, 2022 using the following search terms, " ((((COVID-19 vaccine[MeSH Terms]) AND ((social media[MeSH Terms]) AND (sentiment analysis[MeSH Terms]) AND (demographic[MeSH Terms]))) ) AND (("2019/12/01"[Date - Publication] : "2022/03/01"[Date - Publication]))". The search identified 0 articles. Although previous studies have been published on sentiment analysis towards COVID-19 vaccines, the studies have mostly been survey-based with limited sample size and cohort diversity. Other studies touched upon sentiment analysis using Twitter data. However, recognizing the exorbitant disparities in vaccine uptake to understand hesitancies within specific subpopulations is still a gap. Thus, previous studies have limited potential to guide the promotion of tailored vaccine strategies.

*Added value of this study*

We used deep learning methods to perform sentiment analysis on over 11 million COVID-19 vaccine-related tweets generated on Twitter by over two million unique users. Additionally, we employed validated machine learning methods to retrieve the demographic and geographic information of the users. This enabled us to evaluate vaccine sentiments among multiple subpopulations of users stratified by age, gender, race or ethnicity, and geographic region. Furthermore, since pregnant people were excluded from initial COVID-19 vaccine trials, we conducted a sub-analysis focusing on pregnancy-specific concerns. The trained transfer learning model for sentiment analysis along with the algorithms for demographic, geographic, and pregnancy status information retrieval were integrated into a pipeline that can be implemented for other research projects. The human-annotated Twitter data for finetuning of the sentiment analysis model is also open-sourced to researchers.

*Implications of all the available evidence*

Our findings indicate that the user sentiment towards COVID-19 vaccines became more positive over time, albeit with several periods of fluctuations. The population-level vaccination uptake also increased but lagged the escalation of positive sentiments. Encouraging news and motivating events regarding the development and distribution of vaccines appeared to be associated with the crests in positive sentiment with almost seamless time intervals. Additionally, significant differences in sentiment distributions were found among user subpopulations. Countries and regions around the globe also illustrated diverse patterns in sentiments towards COVID-19 vaccines. The sub-analysis regarding pregnant-specific concerns illustrated delayed patterns in both positive sentiment and vaccine uptake when compared with the general population, suggesting that tailored guidelines or recommendations for populations with extra considerations could alleviate vaccine hesitancy and subsequently improve vaccine uptake. Further feature analysis revealed that concerns in embedded in the negative sentiments centered around the clinical trial validity and logics, unpredictable risks and complications, and the urgency of scientific evidence for particular subpopulations, which suggested essential component to improve in vaccine promotion and education during future pandemic. The approach proposed in this study can be feasibly deployed to similar circumstances to tailor strategies targeting subpopulations for vaccine promotion and to provide early signals for vaccine hesitancy in future pandemics.

Introduction

To fight against the SARS-CoV-2 pandemic, vaccines from multiple pharmaceutical companies began to receive FDA approval at the end of 2020 [1]. As of Summer 2022, more than 12.1 billion doses of vaccines have been administered across over 180 countries or regions[a]. Discussion regarding the vaccines emerged even earlier, especially on social media platforms [2-4]. Individuals hold different attitudes and opinions towards the vaccines and getting vaccinated during such an unprecedented pandemic [5,6]. At the same time, social media, where user-generated content is emphasized, is involved more profoundly in people's lives. Information travels further on the internet and social media, including the news, commentaries, anecdotes, and personal feelings about COVID-19 vaccines.

In previous studies, researchers have attempted to understand attitudes towards COVID-19 vaccination from various perspectives. Some studies administered survey-based methods and associated local distribution strategies focusing on many specific regions [7-9]. Some studies focused on probing specific populations' sentiments towards COVID-19 vaccines[10-12]. However, survey-based studies are frequently subject to limited sample size and time frame. Other groups sought more generalizable perspectives using user-generated data, such as social media[4,13,14]. Compared to surveys, social media analyses have many advantages, including timeliness, easy accessibility, objectivity, diversity, and generalizability. However, one remaining gap underlines understanding the enormous disparities in vaccine sentiments across subgroups. Previous studies were not comprehensive enough in systematically associating large-scaled social media data with user characteristics in an extensive time frame. Thus, comparisons of sentiments across populations were not rigorously evaluated.

In this study, we took advantage of the granularity of the user-generated data on one of the mainstream social media platforms, Twitter, to investigate the sentiments towards COVID-19 vaccines across subpopulations by race or ethnicity, gender, age group, and geographical location. Embracing the diversity of sentiments towards the vaccines, we unraveled how user characteristics interact with vaccine uptake. Pregnant people were excluded from initial COVID-19 vaccine trials, and administering a novel vaccine during pregnancy involves unique considerations. Pregnancy was subsequently found to be a high-risk condition for COVID-19 severity and complications; thus, we further investigated pregnancy-related sentiments as a sub-analysis. Through this study, our objective was to provide advanced informatic strategies for immunization promotion during this pandemic and for future public health emergencies by developing an algorithm for identifying and analyzing vaccine sentiments among subpopulations.

---

[a] https://www.bloomberg.com/graphics/covid-vaccine-tracker-global-distribution/

Methods

Data extraction

We used user-generated posts from a mainstream social media platform, Twitter. The COVID-19-related tweets database was constructed by the Panacea Lab [15]. Tweet objects were retrieved using the Twitter API [a]. Since tweets might be deleted by the user or removed by the platform at any point, the actual tweets extracted may vary. The date range of extraction applied was March 1, 2020 to Mar 1, 2022, which covered the period when the topic of COVID-19 vaccines initially drew public attention to the date when billions of vaccines were administered worldwide. Vaccine-related tweets were identified by regular expressions, with tweets mentioning vaccine distribution filtered out (details in Supplementary material). Tweets analyzed in this study are all original tweets, i.e., retweets are removed. Only the tweets in the English language were considered.

Sentiment analysis

To automatically recognize the sentiment for each tweet, we trained a supervised deep learning classifier based on a subset of 7,700 randomly selected tweets. We followed the sentiment annotation protocol of our prior study[16] (details in Supplementary material). A state-of-the-art deep learning model for sentiment analysis, XLNet [17], was used and finetuned to detect the sentiments for each tweet automatically. Weights are initialized with pre-trained XLNet, which was pre-trained on largescale general domain corpora. Next, we finetuned the model using the general domain sentiment analysis Twitter dataset, SemEval [18], to familiarize the model with Twitter-specific expressions. Finally, a second finetuning and evaluation were carried out. The 7,700 tweets were split into 6,160 (80%) for training, 308 (4%) for validation, and 1232 (16%) for testing. The batch size was 4; the max token length was 256; the learning rate was 3e-5; the model was finetuned for 8 epochs. The test set was held out until the final evaluation. Extensive feature analysis and visualization were further conducted using `BertViz` [19] to illustrate keywords and cue phrases (details in Supplementary material).

Demographics

Three demographic variables, gender, age, and race or ethnicity (we use "race" hereafter for simplicity), were identified by the profile images using the `deepface` package (V 0.0.68) [20], a hybrid face recognition framework wrapping multiple state-of-the-art models. When the profile image was not available for any tweets from a user, we were unable to detect the demographics of such users. Gender in this study was considered dichotomously, i.e., female and male. Race was classified into Asian, Black, Hispanic, and White. Age was further grouped into four groups, 0-19, 20-39, 40-59, and above 60. These classifications were made to reflect distinct social determinants and experiences of each subgroup that may be associated with vaccine attitudes.

---

[a] https://developer.twitter.com/en/docs/twitter-api

Locations

Locations for each tweet were extracted and parsed from the *"location"* attribute in the *"user"* dictionary of the tweet object. The "*location*" attribute is an optional free-texted attribute generated by users, which possesses a great degree of freedom. We used `*mordecai*` package [21] to parse the locations from free-text form to structured geographic information. Vaccine administration data were taken from the *Our World in Data* (OWID) database for COVID-19 vaccine data [22]. It is noteworthy that not all the countries or regions in the dataset are updated daily. The population estimates for calculating per-capita metrics were also obtained from OWID [a]. Countries with a total of fewer than 1,000 tweets were not considered.

Tweet analysis

*Sentiment change*

Sentiment changes are only considered when a user's sentiment switches between positive and negative (i.e., switching to neutral or switching from neutral are not counted) on subsequent tweets, or vice versa. Counts for the number of users with sentiment changes were defined as the number of users who posted a tweet with positive/negative sentiment on the given day whose previous tweet was of negative/positive sentiment.

*Pregnancy-related tweets and data*

Pregnancy-related tweets were identified from the pool of all COVID-19 vaccine-related tweets by regular expression (details in Supplementary material). Weekly data on COVID-19 cases [b] and vaccinations among pregnant people [c] were obtained from the Center for Disease Control and Prevention (CDC). Since pregnancy-related data provided by CDC are only available for the United States, this sub-analysis focuses only on data for the United States.

Statistical methods

All the daily counts were shown as 7-day averages to smooth any fluctuations. Proportion tests were conducted to compare the percentages between groups when appropriate, of which p-values of pairwise tests were adjusted by the False Discovery Rate (FDR) for multiple test correction. Student's t-tests were conducted to compare normally distributed groups. Implementation detail can be found in the Supplementary material.

Results

Data

In total, 11,211,672 vaccine-related tweets corresponding to 2,203,681 unique users from March 1, 2020 to March 1, 2022 were included in the analysis. There were 2,712,310 tweets from

---

[a] https://ourworldindata.org/grapher/population-past-future (Apr 21, 2021)
[b] https://covid.cdc.gov/covid-data-tracker/#pregnant-population (Apr 22, 2022)
[c] https://covid.cdc.gov/covid-data-tracker/#vaccinations-pregnant-women (Apr 22, 2022)

753,998 users with demographic information. For those users, the distribution of demographic features stratified by both the number of users and the number of tweets can be found in *Figure 1*. Distributions by users and by tweets showed similar patterns. The distribution of race was unbalanced, with most of the population being White. The age distribution was centered around approximately 30, with a slight right-skewed pattern and thin tails on both sides. Male users dominated the group by approximately 89%, in line with a recent Statista survey [a] showing that male users have a greater percentage among all users on Twitter.

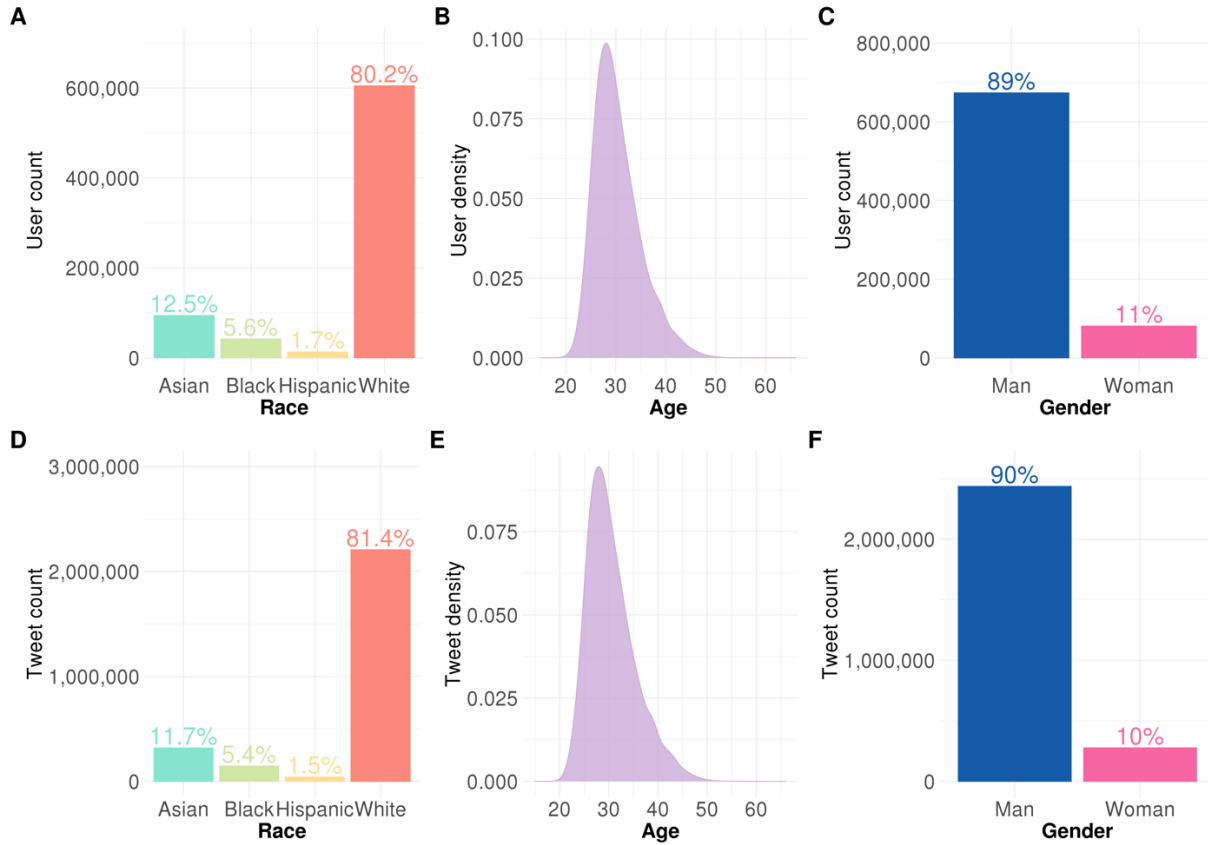

*Figure 1*. Demographic distributions for the number of users (A, B, C) and the number of tweets (D, E, F). **A**. Distribution of race among all users. **B**. Distribution of age of all users. **C**. Distribution of gender of all users. **D**. Distribution of race among all tweets. **E**. Distribution of age among all tweets. **F**. Distribution of gender of all tweets.

Sentiment analysis

The overall accuracy yielded by the finetuned model on the test set of the annotation dataset is 0·92, with a balanced distribution for each class. The classification matrix can be found in *Table 1*.

---

[a] https://www.statista.com/statistics/678794/united-states-twitter-gender-distribution/ (May 18, 2021)

*Table 1.* Performance of sentiment classification by class

|                   | Precision | Recall | F1-score | Support |
|-------------------|-----------|--------|----------|---------|
| negative          | 0·88      | 0·91   | 0·90     | 216     |
| neutral           | 0·94      | 0·90   | 0·92     | 518     |
| positive          | 0·93      | 0·95   | 0·94     | 498     |
|                   |           |        |          |         |
| macro averaged    | 0·92      | 0·92   | 0·92     | 1232    |
| weighted averaged | 0·92      | 0·92   | 0·92     | 1232    |

Tweet trends

*Temporal*

The number of vaccine-related tweets stratified by sentiments is shown in *Figure 2* overlaid with daily vaccination counts. Sentiments set off heterogeneous during the early periods of the studied timeframe. While shortly after the rollout of the vaccine, positive sentiments surpassed neutral and negative sentiments in early January of 2021 (after the dashed green line) and remained dominant for the rest of the studied period. An upswing was also observed in the vaccine uptake but lagged the prevalence of the positive sentiments. Four crests for both positive sentiments and vaccine administration were annotated in *Figure 2*. In addition to the overall trend, the four peaks of positive sentiments also precede each of the peaks in vaccine administrations. Surrounding the peaks of positive sentiments, we detected encouraging news and motivative events happening globally and propagating on social media (*Table 2*). It is noteworthy that numbers in *Table 2* are shown as unaveraged counts of tweets or vaccinations and that the exact numbers and dates can be different from *Figure 2*.

*Fluctuation of sentiments in tweets*

The tendencies of sentiment fluctuation are shown in Figure 2. Two outstanding peaks towards positive sentiments (green) can be observed in the figure. The first peak was on November 9, 2020, when Pfizer-BioNTech announced that the vaccine candidate against COVID-19 achieved success in the first interim analysis from a phase 3 study. The second is on August 23, 2021, when FDA granted full approval to the first COVID-19 vaccine. At the same time, the highest peak towards negative sentiments (red) was on April 13, 2021, when Johnson & Johnson vaccine paused after reports of rare clotting cases emerged.

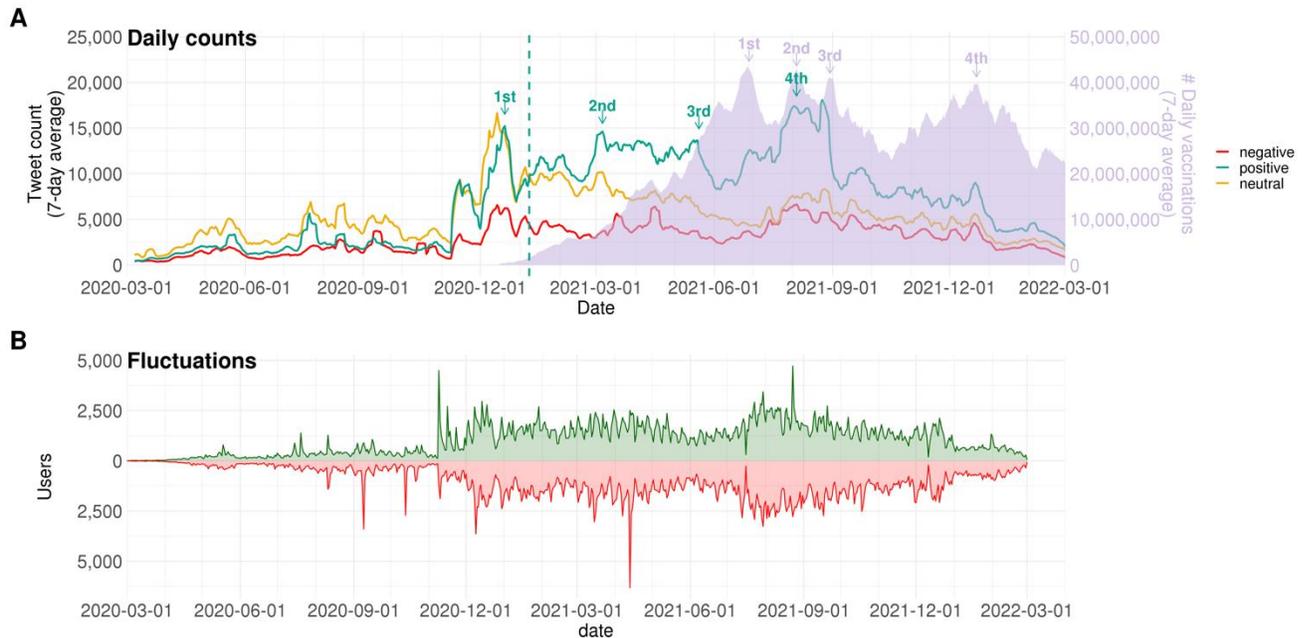

*Figure 2.* Temporal trends of tweets and fluctuations. **A**. 7-day averaged daily counts of tweets (left y-axis) and vaccinations (right y-axis) from March 1, 2020 to March 1, 2022. Daily counts of tweets are shown in negative, positive, and neutral sentiments by color. The number of vaccinations administered daily is illustrated in the purple shade. "1st", "2nd", "3rd", and "4th" denote the four peaks of positive tweets (green) and vaccination counts (purple). The green vertical dashed line marks the time point when positive sentiment started to dominate after largely trending with neutral and negative sentiments. **B**. The fluctuations of sentiments in tweets over time. The number of users who switched to positive sentiments on a day is illustrated in the green area above the axis, while the number of users who switched to negative sentiments on a day is illustrated in the red area underneath the axis.

*Table 2.* Dates with the greatest number of positive tweets around the four peaks

| Peak | Date | Positive Tweet Counts | Events |
|------|------|----------------------|--------|
| 1st  | 2020-12-14 | 20,681 | Sandra Lindsay, a nurse in New York, became the first person in the US to get the COVID-19 shot. |
| 2nd  | 2021-03-02 | 17,788 | Single-dose vaccine from Johnson & Johnson received FDA approval in the US. |
| 3rd  | 2021-05-13 | 20,367 | The Pfizer COVID-19 vaccine was authorized for adolescents 12-15 years old. |
| 4th  | 2021-08-23 | 31,987 | FDA approved first COVID-19 vaccine. |

*Distribution by race or ethnicity, gender, and age group*

The distributions of sentiment towards the vaccines by race are shown in proportions in the top panel of *Figure 3*. Each sentiment panel indicates the percentage of tweets from the given subpopulation. For example, "22%" for the Asian column in the negative subplot can be interpreted as "22% of all tweets posted by Asians individuals were of negative sentiment". The results from the proportion tests are shown by significance level at a type-I error of 0·05 for the comparison among each pair of the subpopulation by race. For the negative tweet proportions, the White subpopulation had slight but significantly fewer negative sentiments compared to the other three racial subpopulations. The distributions of attitudes towards the COVID-19 vaccines by gender are displayed in the middle panel of *Figure 3*. The percentages and the significance levels of the tests can be interpreted similarly to race. While we observed the same proportion of negative tweets in female and male users, female users showed a significantly higher proportion of positive sentiments towards the vaccines. The distributions of attitudes towards the COVID-19 vaccines by age group are shown in the bottom panel of *Figure 3*. The percentages and significance of tests can be interpreted in a similar approach as race and gender. Users aged 20-39 had significantly more positive sentiments towards the vaccine than users aged 40-59. Significant differences in sentiment towards the vaccine can be detected only between the 20-39 and 40-59 age groups, which might be due to the small sample size in younger and senior groups.

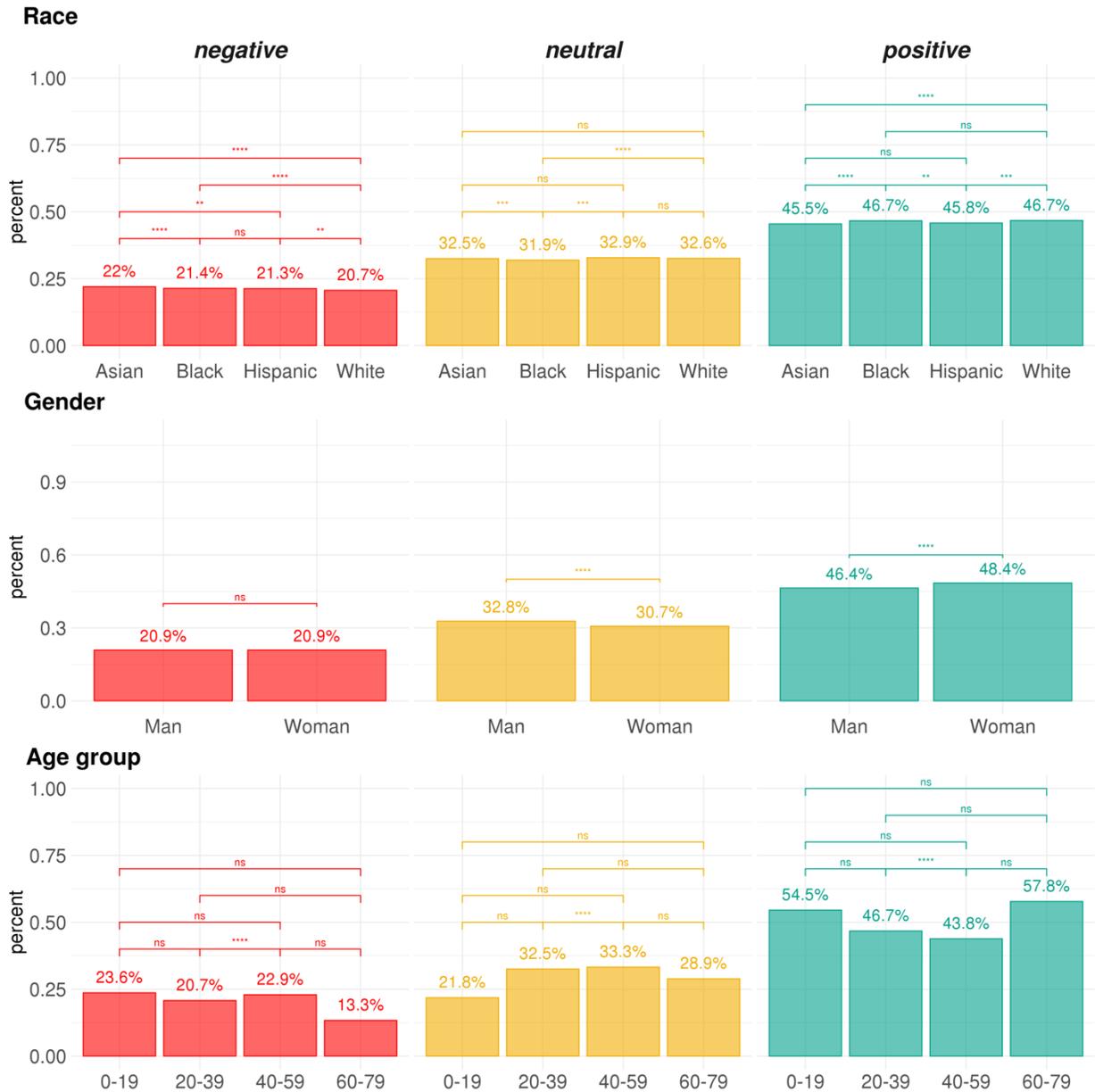

*Figure 3.* Distribution of proportions of sentiments by race, gender and age group. The first column (red) is for negative sentiments; the second column (yellow) is for neutral sentiments; the last column (green) is for positive sentiments. ns: not significant (p-value > 0.05); *: 0.05 < p-value <= 0.01; **: 0.01 < p-value <= 0.001; ***: 0.001 < p-value <= 1e-4; ****: p-value < 1e-4. All the p-values are adjusted for multiple tests using the FDR method.

*Distribution by geographical locations*
Among all 11,211,672 vaccine-related tweets, 6,524,086 tweets had a location parsed from the "*location*" attribute in the "*user*" object. The percentages of positive and negative COVID-19 vaccine-related tweets in each country or region are shown in

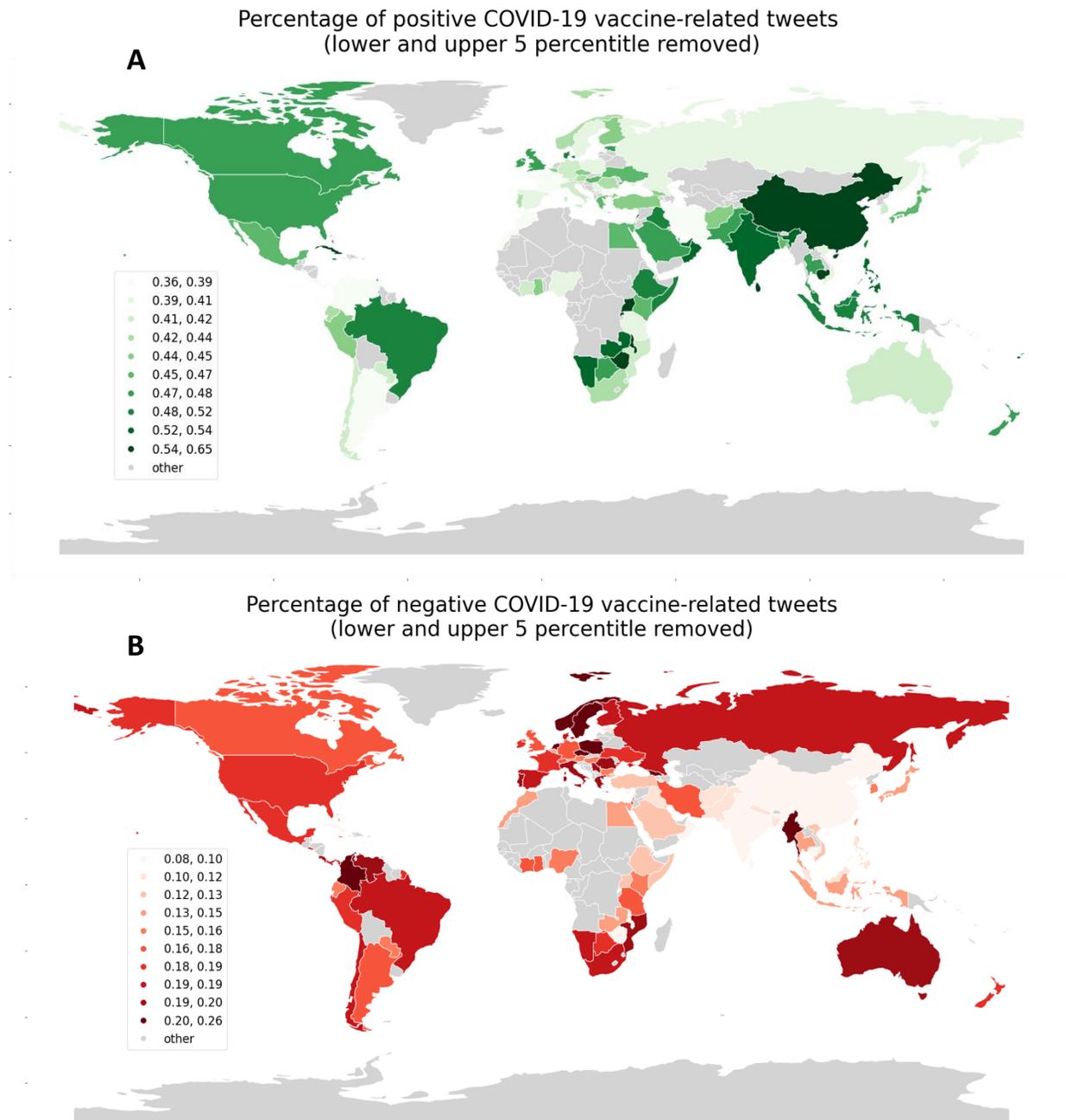

*Figure 4*. The upper and lower five percentiles of the percentages of each sentiment were removed since some countries or regions with a limited number of users may result in extreme percentages. The data are divided into ten quantiles which are illustrated by different shades. Most areas with no data available (gray) are non-English-speaking countries. For regions with data available, some showed a high percentage of positive sentiments and a low percentage of negative sentiments, such as China, India, and Zimbabwe. On the contrary, some regions displayed mostly negative sentiments towards the vaccines, such as Australia, Sweden, and Colombia. Meanwhile, other regions demonstrated controversial attitudes towards the vaccines, such as the United States of America, Canada, and Brazil, where we see equivalent shares in percentages of positive and negative vaccine-related tweets (medium shade in both panels of

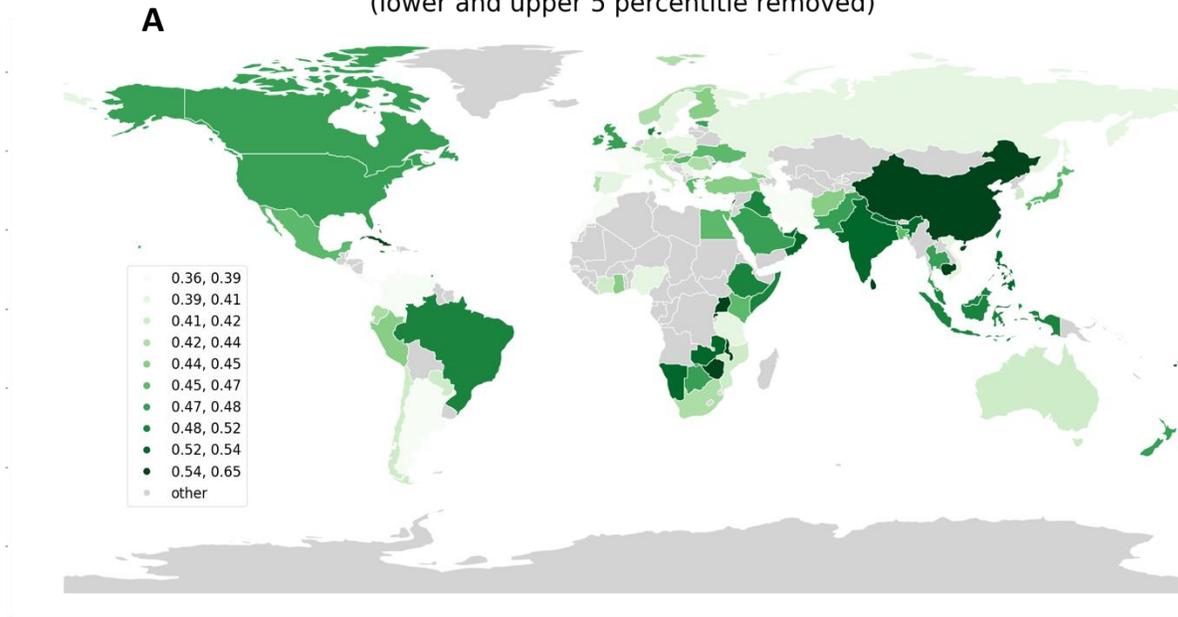

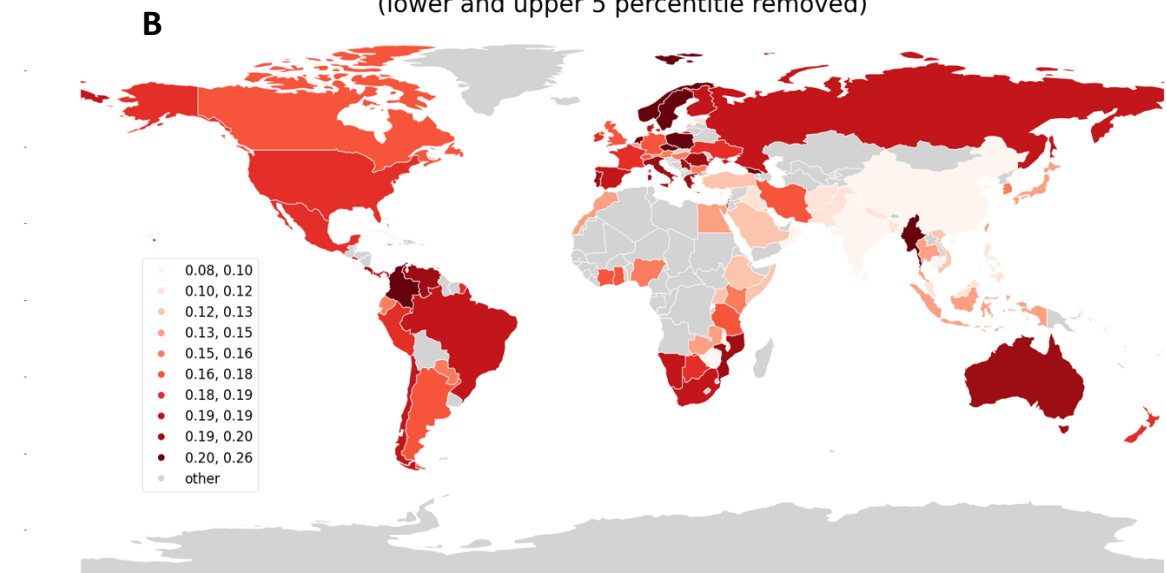

*Figure 4*).

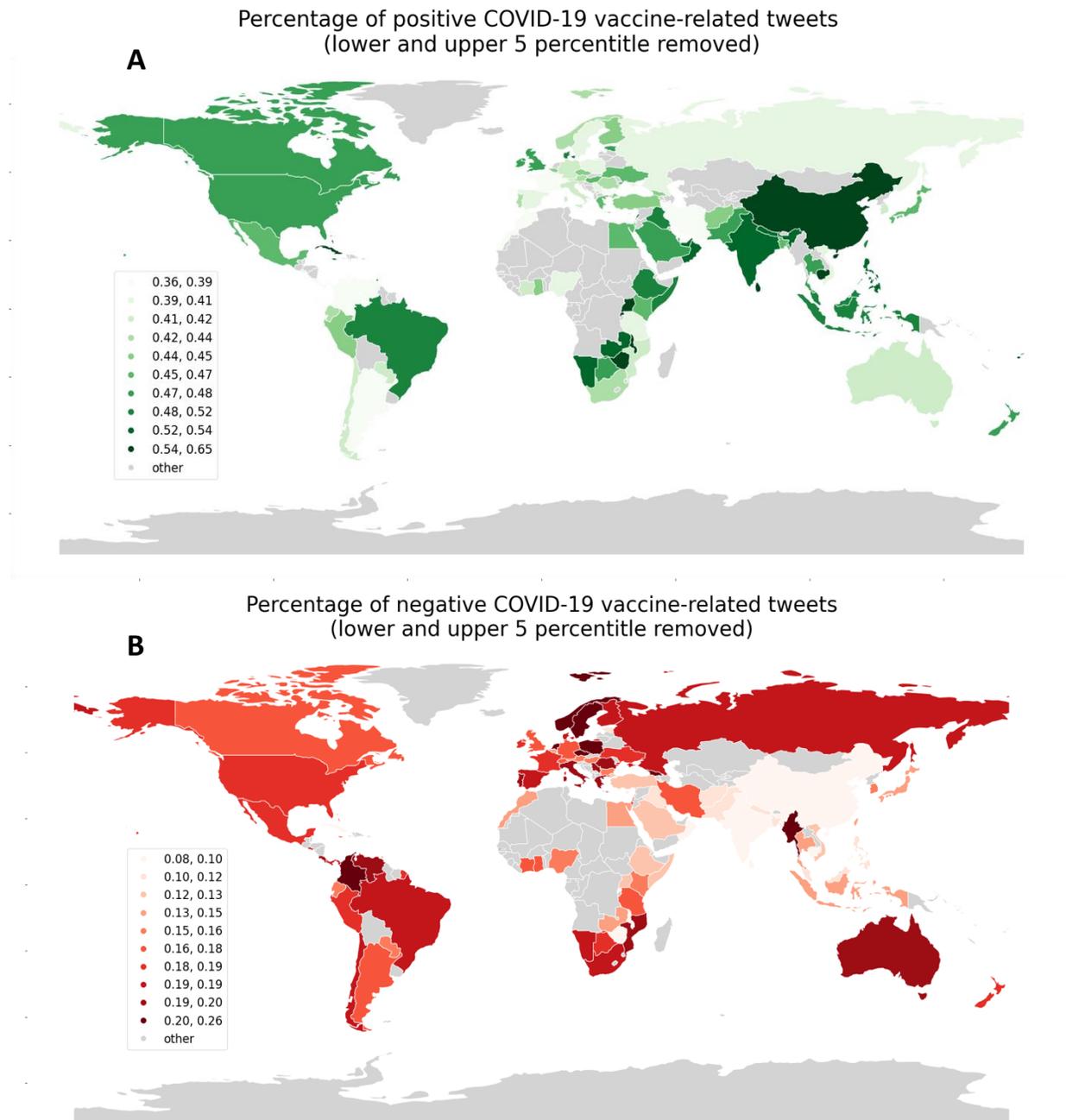

*Figure 4.* Percentages of COVID-19 vaccine-related tweets (with lower and upper five percentiles removed). **A**. Positive tweets; **B**. Negative tweets. Countries or regions excluded or with fewer than 1,000 vaccine-related tweets are shown in gray.

*Pregnancy-related tweets and data compared to non-pregnancy-related tweets and data*
Among all 11,211,672 vaccine-related tweets, we identified 105,039 pregnancy-related tweets, of which 23,325 have a location available in the US. In *Figure 5*, we compared the pregnancy-related trends with the situations in the general population in the United States. Given the initial exclusion of pregnant people in the clinical trials for COVID-19 vaccines, for a period, vaccines were available to pregnant persons without robust. For pregnancy-related sub-analysis, the studied period can be considered as four epochs: 1) pre-evidence-based recommendation stage (until March 7, 2021) [23], 2) efficacy publication stage (until July 29, 2021) [a], 3) recommendation stage (until September 28, 2021) [b], and 4) urgent action stage.

For the general population, the studied period can also be considered as four epochs regarding vaccine development and recommendation: 1) the period when no approved COVID-19 vaccine (until December 11, 2020), 2) the period when vaccines are issued Emergency Use Authorization (EUA) (until August 22, 2021), 3) the period when US Food and Drug Administration (FDA) started to grant full approval to vaccines (until November 29, 2021), and 4) the period when booster shots were recommended by Center for Disease Control and Prevention (CDC).

An overall delayed positive sentiment increase was observed in pregnancy-related tweets. For tweets among the general population, positive sentiments remained dominating once vaccines were granted EUA near the end of 2020. In comparison, the significant peak in positive sentiments among pregnancy-related tweets was around August 11, 2021, when the CDC recommended COVID-19 vaccination for pregnant people based on new safety data [c]. The upswing of vaccination uptake among pregnant people was also delayed when compared with the general population. For the general population, the peak was found in mid-2021, after all three vaccines were granted EUA. In contrast, we observed a gradually increasing trend until early 2022 in the daily vaccination among pregnant people. Furthermore, we can observe two peaks in both populations regarding the trend of COVID-19 cases, with both peaks in pregnant people preceding the general population. The second peak is much higher than the first peak among the general population, while we observed a comparable second peak among pregnant people.

---

[a] https://www.acog.org/news/news-releases/2021/07/acog-smfm-recommend-covid-19-vaccination-for-pregnant-individuals (May 26, 2022)
[b] https://www.cdc.gov/media/releases/2021/s0929-pregnancy-health-advisory.html (May 26, 2022)
[c] https://www.cdc.gov/media/releases/2021/s0811-vaccine-safe-pregnant.html (May 3, 2022)

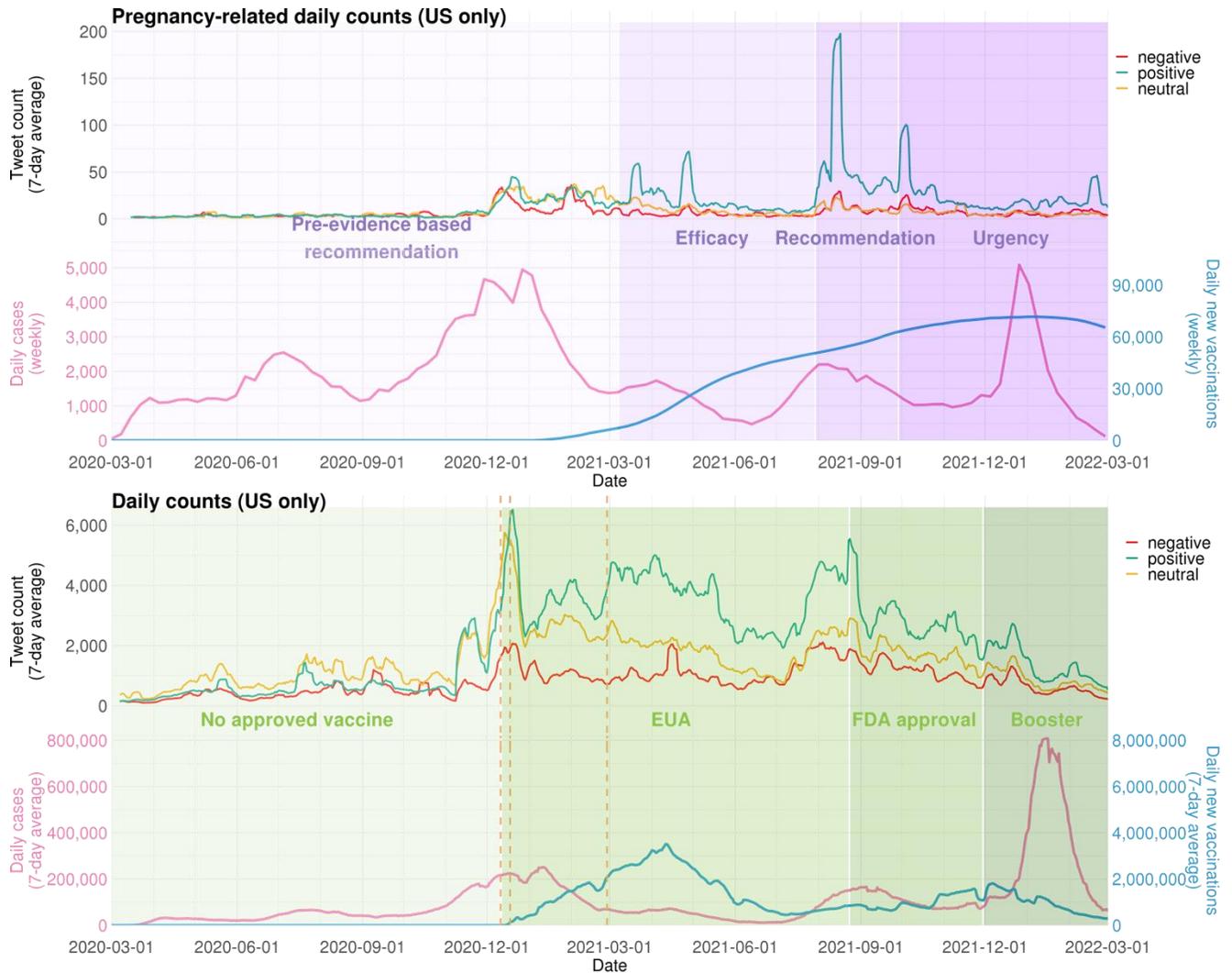

*Figure 5.* Daily pregnancy-related tweet and vaccination and case counts among pregnant women in the United States, compared to the situation among the general population in the United States. Orange vertical dashed lines: EUA dates for Pfizer-BioNTech, Moderna, and Johnson & Johnson COVID-19 vaccines. EUA: Emergency Use Authorization; FDA: US Food and Drug Administration.

Discussion

In this study, we systematically characterized the tendencies of public sentiments towards the COVID-19 vaccines in multiple dimensions using data from a mainstream social media platform, Twitter. Sentiment analyses illustrated distinctive patterns among subpopulations and geographical locations. Sub-analysis of pregnant people revealed more substantial hesitancy compared to the general population.

The XLNet model we used and finetuned to classify sentiment towards the vaccine automatically achieved nearly perfect performance. Transformer models are designed to work across tasks with minimal additional training so that the resulting model can feasibly be re-applied to similar tasks. From March 2020 to March 2022, more than 11 million unique COVID-19 vaccine-related tweets in English were identified. The discussions regarding the vaccine were initiated before the official vaccine rollout, but more tweets were posted in 2021 after clinical trials for the COVID-19 vaccines started showing promising results. Although neutral and negative sentiments continued, positive sentiments have dominated ever since early 2021. The more positive sentiments could demonstrate the success in public health education regarding the vaccine (*Figure 2*). The subsequent upswing in vaccine uptake upon the growth of the positive sentiment implied the significance of vaccination promotion. Meanwhile, the events surrounding the positive sentiment crests suggested vaccination promotion is not limited to slogans and instructions. Instead, developing events that bridge the information gaps, motivating events that share peer experiences, and policies that endorse the vaccine safety and efficacy are all encouraging content that propagates on social media and are associated with positive sentiment upsurges (*Table 2*). Taking advantage of the popularity of social media, vaccine education can be enhanced by stimulating the propagation of positive sentiments by employing supportive content as the medium. Meanwhile, we need more interpretable scientific evidence to manage unnecessary panic for bad news regarding the vaccine, such as the blood clot report of the Johnson & Johnson vaccine.

Analyses of race, gender, and age group revealed distinctive patterns among subpopulations (*Figure 3*). White users were shown to have a lower percentage of negative sentiments. This is in line with a recent study showing that White individuals have lower vaccination hesitancy in multiple countries [24]. Subpopulations of different demographic features might have access to different resources due to the various local policies that might affect the sentiment towards the vaccine. Asian users illustrated higher negative and lower positive sentiments, so we conducted further feature analysis probing the keywords among the negative tweets generated by Asian users (Supplementary Figure S1), which include *"Trial Questioned"*, *"Admits A Mistake (in the trial)"*, *"dangerous"*, *"untested vaccine"*, *"unnecessary"*, *"unethical"*. The keywords suggested concerns regarding validity and rationale in clinical trials, which, in future pandemics, can be alleviated by more explicit clarifications through vaccine education. Male users were found to have significantly higher percentage of negative sentiments. Multiple studies showed higher rates of vaccine side effects among men than women [25,26], which may contribute to the lower number of positive sentiments among male users. Moreover, previous global studies showed that COVID-19 case fatality rates among men are higher than that among women in most countries [27]. Moreover, feature analysis among negative tweets generated by male uses identified following keywords (Supplementary Figure S2), *"fatal"*, *"allergy patients"*, *"related deaths"*, *"blood clotting"*, *"terrible covid vaccine story"*, which revealed concerns towards side effect and complication. Promoting the vaccine based on gender-specific scientific evidence could also elevate the vaccination rate and protect the population. In age group analysis, we only detected significant findings between the 20-39 and 40-59 groups since the 0-19 and 60-79 groups were of small sample sizes. The feature analysis for the 40-59 group, who had high percentage in negative sentiments and low percentage in positive sentiments, probed the following keywords among negative tweets (Supplementary Figure S3), *"sore arm"*, *"Do not trust anything (Fauci says)"*, *"epic failure (around COVID-19 vaccine)"*, *"bad science"*, *"incomplete (covid-19 vaccine study)"*. Bad news and false information hindered the users' trust toward the vaccine, which suggests the necessity of conducting and monitoring proper vaccine promotion. Although not significant, the senior users displayed the lowest negative and highest positive sentiment. Senior citizens are prioritized in vaccine administration in many countries worldwide, which may imply that accessibility and availability of the vaccines also play important roles in shaping users' sentiments.

In the geographical analysis, the patterns of positive and negative sentiments among countries can be described by three categories: high in positive sentiments while low in negative sentiments, low in positive sentiments while high in negative sentiments, and comparable percentages between positive and negative sentiments
(

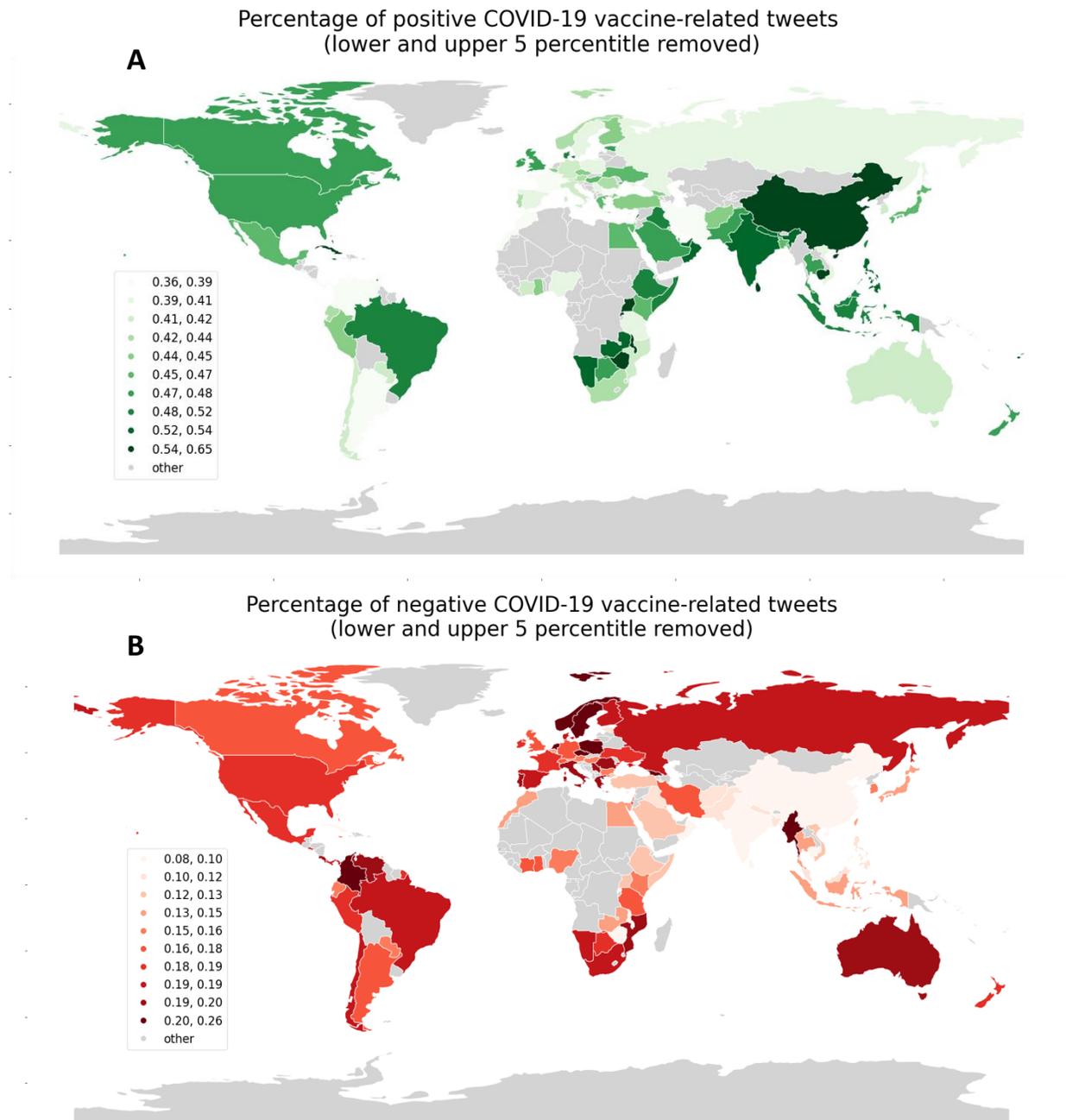

*Figure 4*). Policies regarding COVID-19 vaccinations among countries or regions are different, which come along with various vaccine-promoting strategies. Every country has a unique situation, with distinctive cultures, religions, economic status, and vaccine availabilities, vaccine plans tailored to each exclusive situation would likely improve vaccine uptake in certain places.

Vulnerable populations require extra attention regarding timely evidence-based recommendations and proper education. The pregnant population in this study displayed distinct patterns compared to the general population (*Figure 5*). The two preceding peaks in COVID-19 cases illustrate the escalated susceptibility during pregnancy, which was found to be a high-risk condition for COVID-19 severity and complications [28]. Delay in the vaccination trend upswing illustrated the more substantial hesitancy among pregnant people. Feature analysis probed the keywords and cue phrases in the tweets of negative sentiments (Supplementary Figure S4), which suggested possible evidence for hesitancy. For pregnancy-related analysis, we identified keywords including *"complication associated"*, *"no clinical evidence"*, *"mRNA shots not safe"*; while for the general population in the US, we detected phrases including *"FDA pauses (one of the vaccines)"*, *"risk of worsening clinical disease"*, *"not effective"*. Particularly, *"no clinical evidence"* was highlighted among negative pregnancy-related tweets, which is consistent with the delays that are likely due to the initial trials failed to include pregnant women. A significant surge in positive sentiments was observed associated with the recommendation for pregnant people released on August 11, 2021 based on safety data, demonstrating the significance of tailored guidelines, which also led to an increase in vaccine uptake. Furthermore, the second peak of COVID-19 cases among the general population is considerably higher than the first peak, while the second peak of COVID-19 cases among pregnant people is comparable with the first one, which could benefit from the increased vaccination uptake. If recommendations could have been released earlier for pregnant women, the first peak of the cases among pregnant people could have been lowered. Therefore, timely evidence-based recommendations are essential in protecting vulnerable populations.

In summary, this study revealed the sentiment towards the COVID-19 vaccines on Twitter using a large-scale data set with over 11 million tweets. By analyzing the data in multiple dimensions, we provided suggestions and evidence on vaccine promotion and education. Targeted analyses allow us to focus on vulnerable populations and reveal the importance of timely evidence-based recommendations. We employed a deep learning approach coupled with natural language processing (NLP) techniques which enabled us to extract those features in near-real-time. This approach can be easily deployed for other research applications using Twitter and demographic information as features.

Limitations
We only considered English tweets, which limited coverage in non-English-speaking regions. The algorithms we used to detect demographic variables and geographical locations were adopted from open-sourced repositories [20] which are not guaranteed perfect. Models may need to adapt emerging new words, such as new vaccines or variants of the virus, as the pandemic evolves. Younger and more senior populations were not the predominant users included in our study. Therefore, the findings for those age groups might be weaker than other populations. Additionally, we acknowledge our singular focus on one source of data, Twitter. However, this focused data source helps us to eliminate redundancy since it is highly likely that people have social media accounts across multiple platforms where they may post similar content. Though important for every country, pregnancy-related data are only available in the United States. The data were adopted directly from the CDC website, where we do not have access to raw data for validation. The limitations are all opportunities for future research.

Data sharing

All Twitter data were extracted using the Twitter API [a], and identifiers of COVID-19-related tweets are provided by the Panacea Lab at Georgia State University [15]. Vaccination data [22] and world population data [b] were taken from Our World in Data. Pregnancy-related data were obtained from the Centers for Disease Control and Prevention [cd]. All the data are open-sourced. The annotated data for finetuning the XLNet model can be found on GitHub: https://github.com/luoyuanlab/twitter_vaccine_analysis .

Code availability

The code for the pipeline used to obtain all the variables used in the analysis can be found on GitHub: https://github.com/luoyuanlab/twitter_vaccine_analysis .

Author contribution

Conceptualization of the study: H. Wang, Y. Li, A. Naidech, Y. Luo
Data annotation: H. Wang, Y. Li, M. Hutch, A. Kline
Data analysis: H. Wang, Y. Luo
Paper writing: H. Wang
Pregnancy-related analysis consultation: S. Otero, L. Mithal, E. Miller
Critical revision: H. Wang, M. Hutch, Y. Li, A. Kline, S. Otero, L. Mithal, E. Miller, A. Naidech, Y. Luo

Competing interests

None

Acknowledgments

We thank Northwestern University Quest High Performance Computing for supporting the computation in this study.
Naidech reports funding R01NS110779 and U01NS110772. Luo reports funding R01LM013337 and UL1TR001422. Mithal reports funding NIH/NIAID- K23 AI139337. Miller was site PI for Pfizer phase 2/3 randomized trial of the COVID vaccine in pregnant people.

---

[a] https://developer.twitter.com/en/docs/twitter-api
[b] https://ourworldindata.org/grapher/population-past-future (Apr 21, 2021)
[c] https://covid.cdc.gov/covid-data-tracker/#vaccinations-pregnant-women (Apr 22, 2022)
[d] https://covid.cdc.gov/covid-data-tracker/#vaccinations-pregnant-women (Apr 22, 2022)

Reference

1. Andreadakis Z, Kumar A, Román RG, Tollefsen S, Saville M, Mayhew S. The COVID-19 vaccine development landscape. *Nature reviews Drug discovery* 2020; **19**(5): 305-6.
2. Lyu JC, Le Han E, Luli GK. COVID-19 Vaccine–Related Discussion on Twitter: Topic Modeling and Sentiment Analysis. *Journal of Medical Internet Research* 2021; **23**(6): e24435.
3. Yin F, Wu Z, Xia X, Ji M, Wang Y, Hu Z. Unfolding the determinants of COVID-19 vaccine acceptance in China. *Journal of medical Internet research* 2021; **23**(1): e26089.
4. Puri N, Coomes EA, Haghbayan H, Gunaratne K. Social media and vaccine hesitancy: new updates for the era of COVID-19 and globalized infectious diseases. *Human vaccines & immunotherapeutics* 2020; **16**(11): 2586-93.
5. Roy B, Kumar V, Venkatesh A. Health care workers' reluctance to take the Covid-19 vaccine: a consumer-marketing approach to identifying and overcoming hesitancy. *NEJM Catalyst Innovations in Care Delivery* 2020; **1**(6).
6. Viswanath K, Bekalu M, Dhawan D, Pinnamaneni R, Lang J, McLoud R. Individual and social determinants of COVID-19 vaccine uptake. *BMC Public Health* 2021; **21**(1): 1-10.
7. Largent EA, Persad G, Sangenito S, Glickman A, Boyle C, Emanuel EJ. US public attitudes toward COVID-19 vaccine mandates. *JAMA network open* 2020; **3**(12): e2033324-e.
8. Ward JK, Alleaume C, Peretti-Watel P, et al. The French public's attitudes to a future COVID-19 vaccine: The politicization of a public health issue. *Social science & medicine* 2020; **265**: 113414.
9. Freeman D, Loe BS, Chadwick A, et al. COVID-19 vaccine hesitancy in the UK: the Oxford coronavirus explanations, attitudes, and narratives survey (Oceans) II. *Psychological medicine* 2020: 1-15.
10. Momplaisir F, Haynes N, Nkwihoreze H, Nelson M, Werner RM, Jemmott J. Understanding drivers of COVID-19 vaccine hesitancy among Blacks. *Clinical infectious diseases: an official publication of the Infectious Diseases Society of America* 2021.
11. Harapan H, Wagner AL, Yufika A, et al. Acceptance of a COVID-19 vaccine in Southeast Asia: a cross-sectional study in Indonesia. *Frontiers in public health* 2020; **8**.
12. Bunch L. A tale of two crises: Addressing Covid-19 vaccine hesitancy as promoting racial justice.  HEC forum; 2021: Springer; 2021. p. 143-54.
13. Wilson SL, Wiysonge C. Social media and vaccine hesitancy. *BMJ Global Health* 2020; **5**(10): e004206.
14. Benis A, Seidmann A, Ashkenazi S. Reasons for taking the COVID-19 vaccine by US social media users. *Vaccines* 2021; **9**(4): 315.
15. Banda JM, Tekumalla R, Wang G, et al. A large-scale COVID-19 Twitter chatter dataset for open scientific research—an international collaboration. *Epidemiologia* 2021; **2**(3): 315-24.
16. Wang H, Li Y, Hutch M, Naidech A, Luo Y. Using Tweets to Understand How COVID-19–Related Health Beliefs Are Affected in the Age of Social Media: Twitter Data Analysis Study. *J Med Internet Res* 2021; **23**(2).
17. Yang Z, Dai Z, Yang Y, Carbonell J, Salakhutdinov RR, Le QV. Xlnet: Generalized autoregressive pretraining for language understanding. *Advances in neural information processing systems* 2019; **32**.
18. Rosenthal S, Farra N, Nakov P. SemEval-2017 task 4: Sentiment analysis in Twitter. Proceedings of the 11th international workshop on semantic evaluation (SemEval-2017); 2017; 2017. p. 502-18.

19. Vig J. A multiscale visualization of attention in the transformer model. *arXiv preprint arXiv:190605714* 2019.
20. Serengil SI, Ozpinar A. HyperExtended LightFace: A Facial Attribute Analysis Framework. 2021 International Conference on Engineering and Emerging Technologies (ICEET); 2021: IEEE; 2021. p. 1-4.
21. Halterman A. Mordecai: Full Text Geoparsing and Event Geocoding. *J Open Source Softw* 2017; **2**(9): 91.
22. Mathieu E, Ritchie H, Ortiz-Ospina E, et al. A global database of COVID-19 vaccinations. *Nature human behaviour* 2021: 1-7.
23. Gill L, Jones CW. Severe acute respiratory syndrome coronavirus 2 (SARS-CoV-2) antibodies in neonatal cord blood after vaccination in pregnancy. *Obstetrics & Gynecology* 2021; **137**(5): 894-6.
24. Nguyen LH, Joshi AD, Drew DA, et al. Self-reported COVID-19 vaccine hesitancy and uptake among participants from different racial and ethnic groups in the United States and United Kingdom. *Nature communications* 2022; **13**(1): 1-9.
25. Saeed BQ, Al-Shahrabi R, Alhaj SS, Alkokhardi ZM, Adrees AO. Side effects and perceptions following Sinopharm COVID-19 vaccination. *International Journal of Infectious Diseases* 2021; **111**: 219-26.
26. Abu-Hammad O, Alduraidi H, Abu-Hammad S, et al. Side effects reported by Jordanian healthcare workers who received COVID-19 vaccines. *Vaccines* 2021; **9**(6): 577.
27. Dehingia N, Raj A. Sex differences in COVID-19 case fatality: do we know enough? *The Lancet Global Health* 2021; **9**(1): e14-e5.
28. Wastnedge EA, Reynolds RM, Van Boeckel SR, et al. Pregnancy and COVID-19. *Physiological reviews* 2021; **101**(1): 303-18.